# Persona-based Multi-Agent Collaboration for Brainstorming

Nate Straub*
*Reality Labs*
*Meta*
Menlo Park, CA, USA
nstraub@meta.com
*co-first author

Saara Khan*
*Reality Labs*
*Meta*
Menlo Park, CA, USA
saarak@meta.com
*co-first author

Katharina Jay
*Reality Labs*
*Meta*
Menlo Park, CA, USA
kattp@meta.com

Brian Cabral
*Reality Labs*
*Meta*
Menlo Park, CA, USA
bkc@meta.com

Oskar Linde
*Reality Labs*
*Meta*
Menlo Park, CA, USA
oskar@meta.com

*Abstract*—We demonstrate the importance of persona-based multi-agents brainstorming for both diverse topics and subject matter ideation. Prior work has shown that generalized multi-agent collaboration often provides better reasoning than a single agent alone [1]. In this paper, we propose and develop a framework for persona-based agent selection, showing how persona domain curation can improve brainstorming outcomes. Using multiple experimental setups, we evaluate brainstorming outputs across different persona pairings (e.g., Doctor vs VR Engineer) and A2A (agent-to-agent) dynamics (separate, together, separate-then-together). Our results show that (1) persona choice shapes idea domains, (2) collaboration mode shifts diversity of idea generation, and (3) multi-agent persona-driven brainstorming produces idea depth and cross-domain coverage.

## I. INTRODUCTION

Brainstorming has historically been a human-centered activity where diverse individuals bring unique knowledge and perspectives to generate novel ideas. Locke's theory of knowledge formation emphasizes that combining and abstracting experiences across multiple people leads to more complex ideas. Similarly, since the 1950s and '60s, design thinking frameworks emphasize the importance of multiple participants generating and refining ideas through structured exploration of brainstorming to generate ideas for a pre-defined question [2]. These design thinking frameworks use a set of cognitive, strategic, and practical procedures for ideation [2] and for this paper we focus on 'brainstorming' as an area of exploration for multi-agent collaboration. Brainstorming is normally done with multiple and diverse humans standing at a whiteboard together brainstorming ideas against a topic area that is put on the whiteboard. By practice, the people selected to brainstorm together should be from different fields for the best cross-pollination of ideation and domain experts in various fields are encouraged to come together. The goal of brainstorming is to come up with as many ideas as possible before sorting, organizing, and down-selecting one idea that should be prototyped. Our goal in this paper is to explore if we can develop multi-agents to generate a diverse set of ideas across specified domains with multi-agents to get us closer to human-centered reasoning for brainstorming.

With the advent of large language models (LLMs), multi-agent systems have emerged as a promising tool for ideation. Single-agent prompting often yields repetitive or limited outputs, while multi-agents setups allow for debate, critique, and cross-pollination of ideas. Within single agent development, chain-of-thought has drastically improved reasoning abilities [3], where reasoning steps are provided in few-shot examples, leading to a substantial enhancement in reasoning capabilities of LLMs. To further reasoning capabilities, multi-agent systems are under investigation and have shown to further enhance capabilities as well so that reasoning isn't happening within one agent, but across agents [1]. However, most multi-agent literature has focused on the agent-to-agent interaction with the agents being generalized or given soft personas related to their thinking style [4]. In addition, in literature persona-based research has started to show that personas can improve reasoning compared to their generic counterparts [5], but this has not been directly applied to multi-agent brainstorming.

Before diving into multi-agent collaboration, it is critical to take a step back and ask a foundational question: what kinds of agents should we be putting together in the first place? While most work in this space focuses on agent-to-agent interaction dynamics, this paper takes a principled approach inspired by human collaboration—where the success of brainstorming depends heavily on who is in the room and the mix of their expertise, perspectives, and reasoning styles. Translating that to AI systems, we hypothesize that the persona and role definition of each agent fundamentally shape the creativity, diversity, and coherence of the ideas that emerge during multi-agent brainstorming. By systematically studying how different persona pairings (e.g., domain expert versus generalist, complementary versus redundant expertise) interact across varied collaboration modes—such as separate

This work was supported by funding from Meta Platforms.

ideation, joint ideation, or hybrid configurations—this work aims to provide a framework for principled agent selection. The goal is to guide future system design so that AI teams are intentionally composed, not random—bridging human-centered design principles with the next generation of multi-agent AI systems.

In this paper, **our findings show that the personas and the brainstorming dynamics chosen for the agents fundamentally shape the brainstorming landscape**. We demonstrate these results through different persona pairings (e.g., Doctor vs VR Engineer) and A2A dynamics (separate, together, separate-then-together). We built an A2A system that is able to generate these ideas in a similar UI to what humans would traditionally write down on a whiteboard to generate these ideas and allow the user to pick the personas and define the prompt for the agents.

The main contributions of this paper are as follows:
- **Multi-Agent Brainstorming System:** An A2A framework and system for persona-based multi-agent brainstorming that ties prompt → domain breakdown → agent selection → interaction style → output.
- **Generalist vs Persona Results**: Experimental results comparing generalist vs. persona-driven agents, demonstrating clear shifts in idea coverage and entropy.
- **Innovative A2A Collaboration Dynamics**: An empirical analysis of collaboration dynamics (separate, together, separate-then-together) and their impact on ideation
- **Comprehensive Evaluation**: Visualization methods (PCA, entropy, thematic progression) for analyzing idea diversity in multi-agent brainstorming.

## II. RELATED WORK

### A. Human Centered Design and Brainstorming

Brainstorming has historically been a human-centered, collaborative activity as a formal creativity technique for idea generation [6]. Design thinking frameworks incorporate brainstorming as a key ideation phase, emphasizing divergent thinking (generating many ideas) followed by convergent thinking (refining and selecting ideas) in multidisciplinary teams [6]. The underlying rationale is that teams of individuals with diverse knowledge and perspectives can cross-pollinate ideas and produce more diverse solutions than any single person alone [7]. Research in creativity supports this: heterogeneous groups tackling complex problems tend to yield more diverse ideas, as different members contribute unique viewpoints that collectively expand the idea space [7]. In other words, flexibility and domain knowledge "need not be present in the same individual... as long as they are present within the group," enabling more divergent and innovative outcomes [7]. Modern human-centered design practice therefore encourages collaborative brainstorming sessions where participants suspend early judgment, build on each other's ideas, and leverage their varied expertise to drive creative ideation [6]. Our work is inspired by these principles – we translate the multi-perspective brainstorming ethos from human design teams into a multi-agent context, using multiple persona-based agents to mimic the diversity and cognitive synergy of human collaborators.

### B. Multi-Agent Collaboration in LLMs

Large language models have demonstrated improved reasoning ability when employing structured prompt strategies like chain-of-thought, which guides a single agent to think step-by-step [8]. However, a single LLM reasoning in isolation can struggle to spot its own blind spots or errors [8]. This limitation has spurred growing interest in multi-agent collaboration frameworks, where multiple LLM-based agents interact to solve problems collectively. Inspired by Minsky's "Society of Mind" idea that intelligence can emerge from multiple interacting experts [9], researchers have developed multi-agent discussion and debate techniques to enhance reasoning. In these systems, two or more agents exchange their reasoning, critique each other's answers, and refine solutions through dialogue [8]. For example, the Multi-Agent Debate (MAD) framework recruits several LLM agents to work on the same question, share their intermediate answers, identify disagreements, and then arrive at a final solution via a form of majority voting [8]. Such multi-agent discussions are reported to outperform a lone chain-of-thought reasoner on many tasks, essentially because agents can catch each other's mistakes and combine their knowledge [9]. Indeed, numerous studies claim that involving multiple specialized agents leads to better reasoning accuracy and problem-solving performance than single-agent prompting methods [8]. These advances show that collaborative reasoning – analogous to a team of experts deliberating – can push beyond the capability of one model working alone. Most existing multi-agent LLM approaches, however, use agents with either identical general-purpose roles or predefined debate roles (e.g. critic and solver). In contrast, our work explores a different dimension: equipping each agent with a distinct persona or domain expertise during collaboration. By doing so, we aim to simulate the effect of an interdisciplinary team of AI specialists brainstorming together, which has not been a focus of prior multi-agent reasoning research. We build on the multi-agent paradigm but differentiate our approach by systematically studying how persona diversity and different collaboration structures influence the breadth and depth of generated ideas.

### C. Persona Use in AI

Leveraging personas or roles in AI systems has emerged as a powerful prompt engineering technique to guide generation and encourage diverse outputs. Recent studies have shown that prompting an LLM with a specific role-playing persona can significantly improve its reasoning and problem-solving capabilities in zero-shot settings [10]. For instance, assigning a model the persona of a domain expert (e.g. "You are a math teacher") tends to elicit more focused, step-by-step solutions in that domain than a neutral prompt [5]. In general, persona-conditioned prompting leads to more in-character and contextually tailored responses. Commercial systems have embraced this idea: platforms like Character.AI deploy dialogue agents with a wide range of personalities, demonstrating that role-play can yield varied conversational styles and insights [5]. On the analytical side, persona prompts can act as an implicit trigger for logical reasoning – effectively inducing the model to perform chain-of-thought reasoning under the guise of a character's thinking process. Empirical evaluations on numerous NLP benchmarks confirm that carefully chosen personas often boost accuracy and richness of outputs relative to standard prompts [5]. However, most prior work on persona-based AI has focused on single-agent use cases: a lone model adopting a role to enhance its performance or creativity. Common applications include persona-consistent dialogue generation, character-driven storytelling, and role-based question answering, typically using one agent conditioned on a given profile. In contrast, our research applies personas in a multi-agent brainstorming methodology. By instantiating multiple agents with different personas and having them jointly generate and refine ideas, we explore whether persona diversity can similarly diversify the solution space and improve outcomes in collaborative brainstorming. This approach aligns with the concept of "role-storming" in human creativity, where team members ideate from the perspective of various characters to unlock new angles. We hypothesize – and our results confirm – that persona-driven multi-agent generation produces more diverse and comprehensive idea sets, as each AI agent contributes unique domain insights and styles. Our work therefore extends persona-based prompting beyond single-agent scenarios, demonstrating its benefits in coordinating a team of complementary AI personas toward a creative task.

## III. Core System Framework

To systematically investigate how persona selection and collaboration dynamics influence multi-agent ideation, we developed an experimental framework that enables controlled manipulation of agent roles, interaction patterns (based on human-centered design principles), and cross-turn information access. The system is built on Pydantic AI agents [11], exposed via FastA2A, an implementation of Google's A2A protocol [12], enabling standardized agent-to-agent communication. The experiments are configured and outputs are visualized through a modular front-end user interface, while the session itself is orchestrated via the Python code running on a server once a request is received.

### A. User Interface and Interaction Console

The front-end (Fig. 1) is implemented as a browser-based application that serves as both a configuration console and real-time visualization dashboard. The interface provides three primary interaction surfaces:

- **Persona Selection Grid:** Users select exactly two personas from nine available domain experts (Doctor, Nurse, Dentist, VR Engineer, iOS Engineer, Mobile Engineer, Design Prototyper, UX Researcher, Frontend Designer). Each persona can be assigned custom system prompts that override default role descriptions.
- **Configuration Control Panel:** Exposes three experimental variables:
  - *Interaction Type:* Collaborative
  - *Ideation System:* Three collaboration modes (Separate, Together, or Separate-Then-Together)
  - *Phase Configuration:* Turn count limits for separate ideation and collaborative discussion phases
- **Visualization Canvas:** A whiteboard-style canvas that displays agent-generated ideas as color-coded "sticky notes". Each note has a color mapped deterministically to a persona type (e.g., Doctor–blue, Nurse–yellow). During Separate-Then-Together sessions, the collaborative phase introduces visually distinct combined colors (e.g., blue+yellow=green) to signal cross-pollination. Real-time updates are streamed via WebSocket connections due to the distributed agent design.

User configurations are captured as **SessionConfig** objects containing topic text, persona selections, interaction parameters, and optional custom system prompts, which are transmitted to the backend via REST API for session instantiation.

### B. Session Engine & Distributed Agent Runtime

The back-end consists of two main components: a Session Engine and a Distributed Agent Runtime.

The **Session Engine** serves as the central orchestration component, integrating conversation strategy management with session execution. The engine manages session lifecycle, agent turn execution, and event propagation while enforcing collaboration-mode-specific rules through polymorphic strategy implementations.

Upon session creation, the engine generates a unique session identifier, persists the session record to storage, and instantiates the appropriate conversation strategy based on the selected ideation system. The strategy constructs the phase sequence, which is stored with the first phase activated.

A session loop then executes the following control flow until completion:

1) Check phase completion status and evaluate transition conditions
2) If phase is complete and additional phases exist, advance to the next phase
3) Retrieve full conversation history from persistent storage
4) Determine which persona should act next (alternating turns)
5) Build execution context with strategy-specific history filtering applied
6) Execute agent turn through the A2A communication layer
7) Process returned action, assign spatial coordinates for sticky notes
8) Persist action to database and broadcast to connected clients
9) Increment phase turn counter

All session states, agent actions, sticky notes, and system configuration are persisted to a relational SQLite database through a storage interface. The schema includes tables for sessions, actions, sticky notes, prompts, and configuration, with indexed lookups on session identifiers and persona fields to support efficient query patterns during history retrieval.

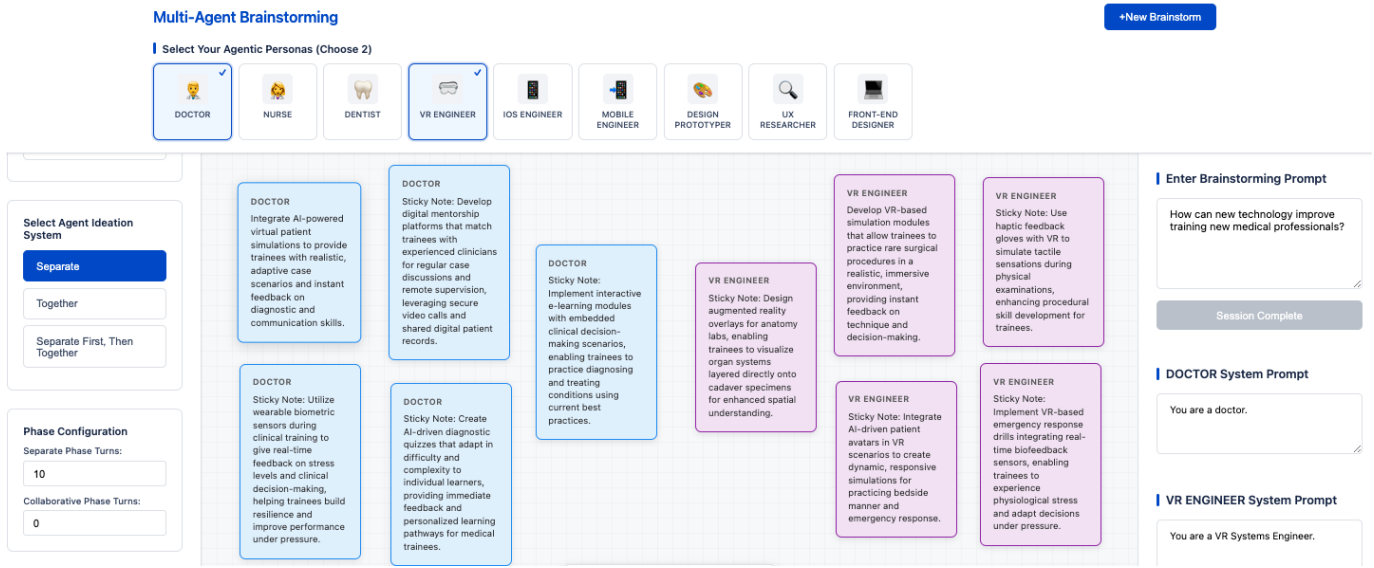

Fig. 1. The front-end UI for agent-based multi-agent brainstorming is implemented as a browser-based application that serves as both a configuration console and real-time visualization dashboard.

The **Distributed Agent Runtime** consists of Pydantic AI Agent objects mounted as FastA2A applications on a FastAPI server. During server initialization, the system retrieves the current model configuration (model name and temperature) from storage for each persona and instantiates the appropriate model provider. It then converts it to a FastA2A application, and mounts it on the persona-specific path. This mounting process creates isolated agent instances that independently handle A2A protocol requests without shared state, enabling concurrent turn processing if desired. Communication is mediated through an A2A client that implements Google's A2A protocol using JSON-RPC 2.0 over HTTP. The protocol provides a standardized interface for LLM-to-LLM communication, enabling multi-agent orchestration.

For each agent turn, the client constructs a structured user prompt that includes the brainstorming topic and current phase name, turn constraints, partner agent identification, filtered conversation history (strategy-dependent: own actions only vs. full dialogue), and phase-specific instructions (independent ideation vs. collaborative discussion). Messages are encapsulated in JSON-RPC payloads and sent to agent endpoints, which process messages asynchronously and return task identifiers. The client polls for task completion at 1-second intervals until the agent's response is ready. The response text is converted into an AgentAction object with session metadata attached. These real-time updates are then propagated to connected clients via a WebSocket manager that maintains a registry of active connections per session. When agents produce actions, phase transitions occur, or sessions complete, the manager serializes events as JSON and broadcasts to all clients subscribed to the relevant session.

The server also implements a dynamic reconfiguration mechanism where changes to model selection or temperature trigger a coordinated remount sequence—the system acquires a lock to prevent concurrent remounts, persists the new configuration, signals a background agent manager to initiate remount, performs asynchronous cleanup of all mounted agents, unmounts existing routes, and re-executes the mounting process with the new configuration. This architecture allows mid-experiment model switching without server restart, facilitating comparative experiments across different language model families.

### C. Multi-Agent A2A Dynamics

To accommodate the three ideation systems under investigation, the engine employs distinct conversation strategies that encapsulate turn-taking logic, context filtering, and phase transition rules (see Fig. 1):

1) **Separate Strategy:** Implements independent ideation where agents work in parallel without mutual awareness. Initializes a single separate ideation phase with discussion disabled. History filtering includes only the current agent's prior actions, ensuring epistemic isolation. Turn-taking alternates deterministically between the two personas. For our experiment, this was set to 30 (15 ideas per agent).

2) **Collaborative Strategy:** Implements joint ideation with full conversation visibility. Initializes a single collaborative brainstorm phase with discussion enabled. Context building provides unfiltered conversation history, allowing agents to reference, critique, and build upon partner contributions while maintaining alternating turns for balanced participation. For our experiments, this was set to 30 (15 ideas per agent).

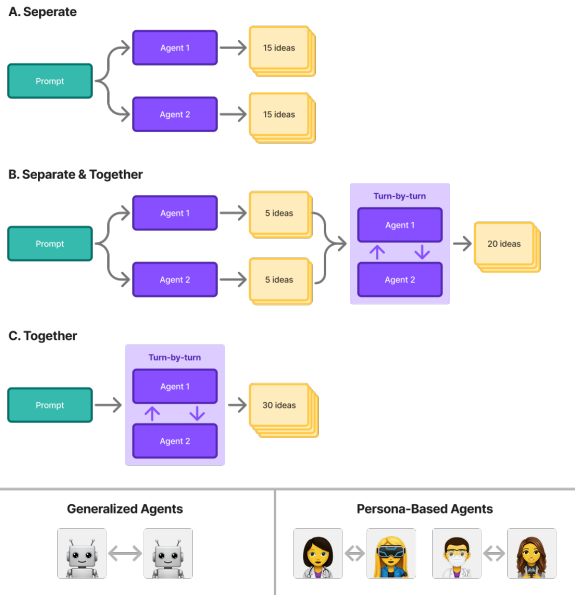

Fig. 2. Brainstorming dynamics built into our system for various phase configurations for multi-agent brainstorming. For these three ideation systems, the engine employs distinct conversation strategies that encapsulate turn-taking logic, context filtering, and phase transition rules.

3) **Separate-Then-Together Strategy:** Implements a two-phase hybrid approach combining both prior strategies. Initializes two sequential phases: separate ideation followed by collaborative discussion. During the separate phase, employs history filtering to isolate agents; during the collaborative phase, switches to full history access. Phase advancement triggers automatically when turn limits are reached, enabling the system to shift from isolated ideation to collaborative refinement. For our experiment, we set Separate Phase turns of 10 (5 ideas per agent), and then collaborative phase turns of 20 (10 ideas per agent).

*D. Persona Selection for Brainstorming*

For brainstorming principles, it is recommended to find brainstorming partners that have a different knowledge base or understanding of the solution space that is different from their own. This allows the collaboration to be fruitful, avoiding groupthink, and generating ideas across more domains [2].

To apply this same principle to AI Agents, we develop system prompts of different agents that are substantially different from one another, but also have the right knowledge base to answer the question that initially prompted them with. We do not tackle trying to automate generating the persona's system prompt based on the question in this paper, but focus our attention on helping humans to create personas and getting a diversity score on how similar or different the personas are from one another. There has been some work around optimizing multi-agents with better prompts and topologies [13], but persona based optimization for multi-agents continues be a fruitful area of research to automate in the future.

Below we developed two different personas that are targeted to answer the question later in the paper **"How can new technology improve training new medical professionals?"** We develop a persona for a healthcare professional (doctor) and develop a persona for a VR Systems Engineer (engineer). To assess the similarity of the persona embeddings we employed cosine similarity as seen in Fig 3. Analyzing the personality of agents using these methods has been done previously by [14], and [15] but were employed for creating AI-to-Human interactions with different personality traits or evaluating the correctness of an agent based on the system prompt. We employ this method to a different application space to create two different agent personas to explore a knowledge space together with brainstorming.

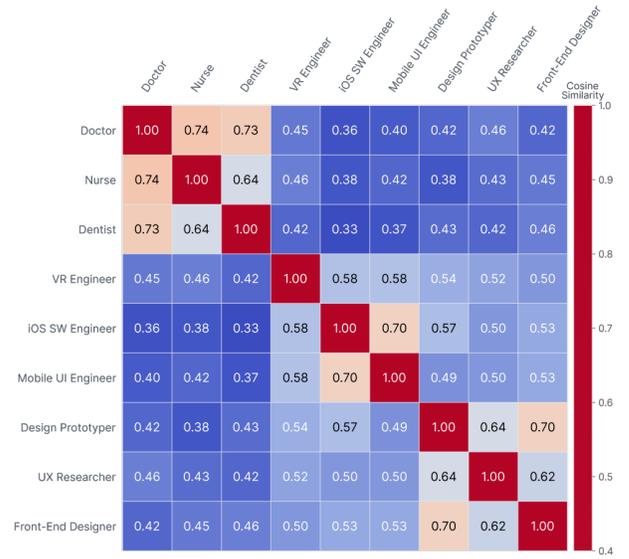

Fig. 3. Cosine similarity between personas for brainstorming, agents are selected to be different or similar based on this matrix.

## IV. RESULTS AND EVALUATION

We evaluated persona effects (general versus persona given) and A2A dynamics for our multi-agent brainstorming system against the question "How can new technology improve training new medical professionals?" Our evaluation metrics focused on PCA plots to visualize clustering and shifts between agents, theme categorization of ideas (domain buckets), and entropy across themes (diversity of coverage). On the backend, we utilized gpt-4.1 with a temperature of 1. We assessed the system across multiple LLMs later on in the evaluation, the model mostly influences the topics generated for our evaluation. Our results show that (1) persona choice shapes idea domains, (2) collaboration mode shifts diversity and entropy of idea generation, and (3) multi-agent persona-driven brainstorming produces subject matter depth and cross-domain coverage.

*A. A2A Persona Brainstorming Behavior*

We utilized Principal Component Analysis (PCA) to visualize the LLM-generated text embeddings from the brainstorm-ing output from multi-agents. Although LLM text embeddings are dispersed across many dimensions with less than thirty percent combined variance for PCA1 and PCA2, it is common to use the first two PCAs to uncover structure and behavior in the data [21]. For text embeddings, PCA1 and PCA2 often align with latent factors such as topic categories and we plotted each dot from the embedding on the graphs.

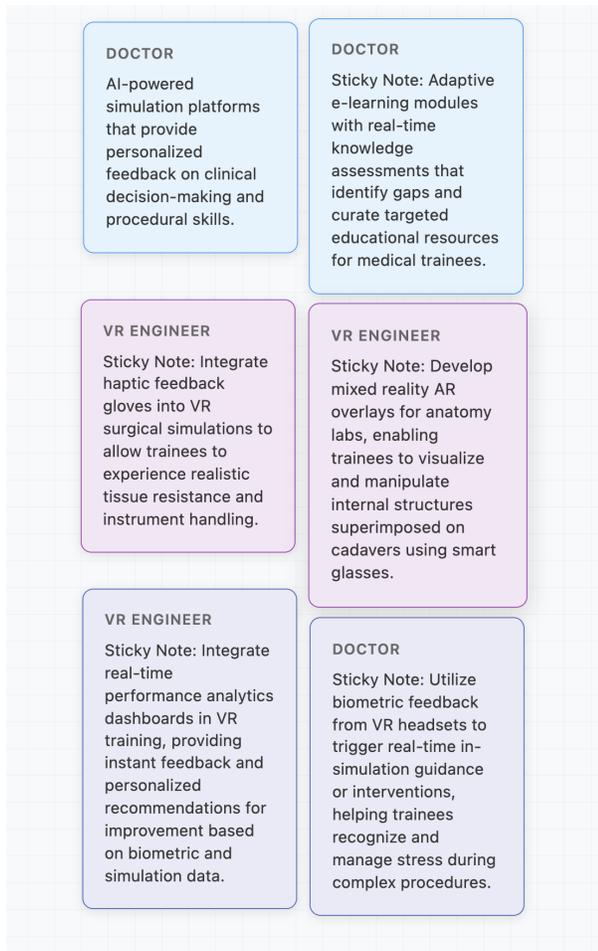

Fig. 4. Brainstorming example output from Doctor and VR Engineer brainstorming is separate then together. Blue sticky notes are Doctor only, pink sticky notes are VR engineer only, and Purple color is from output of brainstorming collaboratively.

All brainstorming transcripts were projected into a unified PCA space to examine semantic diversity and orthogonality across three interaction modes: (A) Separate, (B) Together, and (C) Separate-then-Together. Persona pairs were chosen to represent varying domain similarity: Doctor × VR Engineer (cosine similarity 0.45), Dentist × iOS Engineer (0.33), and Doctor × Nurse (0.74). Two generalist agents without domain conditioning served as the baseline.

For the Generalist Agent pair (control group), PCA projections (Fig. 5, top row) exhibit substantial overlap and low semantic differentiation across all modes.

- **(A) Generalist Pair, Separate:** Idea clusters coincide, showing minimal separation along the first two components. Both agents generate generic and surface-level technological concepts with limited lexical diversity.
- **(B) Generalist Pair, Together:** Clusters intertwine again, suggesting modest cross-influence but no emergence of distinct conceptual perspectives.
- **(C) Generalist, Separate-then-Together:** Clusters continue to intertwine in the separate phase and reconverge toward the PCA center during collaboration, indicating semantic homogenization.

These observations imply that unconditioned agents tend to converge toward similar lexical and conceptual regions.

The Doctor × VR Engineer (Dissimilar Personas) pair demonstrates clear domain-driven separation and directional complementarity (Fig. 5, bottom row).

- **(A) Doctor × VR Eng, Separate:** The Doctor cluster (red) anchors the left PCA region, characterized by clinically grounded and patient-oriented semantics, while the VR Engineer cluster (blue) occupies the right region emphasizing simulation, haptics, and system mechanics.
- **(B) Doctor × VR Eng, Together:** Overlap emerges as both personas integrate each other's vocabulary, showing conceptual convergence and interdisciplinary blending.
- **(C) Doctor × VR Eng, Separate-then-Together:** Initially distant clusters migrate toward the PCA center, achieving balanced integration between clinical and technical domains and exhibiting the highest degree of semantic overlap and idea symmetry.

Dissimilar persona pairs results imply cross-domain synthesis—introducing orthogonal idea spaces that gradually align through collaboration.

The Doctor × Nurse (Similar Personas) visualizes brainstorming between highly similar medical personas (Fig. 6, top row). Their shared expertise produces compact clustering and constrained variance.

- **(A) Doctor × Nurse, Separate:** Doctor and Nurse idea clusters overlap substantially, reflecting convergent clinical reasoning and overlapping conceptual lexicons.
- **(B) Doctor × Nurse, Together:** Overlap between idea cluster continues, with similar PCA distributions, indicating reinforcement of existing conceptual schemas rather than expansion.
- **(C) Doctor × Nurse, Separate-then-Together:** A slight initial dispersion collapses rapidly toward the PCA center, yielding cluster convergence and limited orthogonality.

These results suggest that persona similarity has similar behavior to the control state (generalist agents) by minimizing semantic spread between the two agents. We later will see differences in the themes compared to the control dataset.

The Dentist × iOS Engineer pair (Fig. 6, bottom row) provides a larger contrast in disciplinary framing, yielding in wider PCA dispersion and highest conceptual diversity.

- **(A) Dentist × iOS Eng, Separate:** Distinct clusters emerge—Dentist (red) emphasizing anatomy, health, and procedural reasoning, and iOS Engineer (blue) emphasizing app architecture, sensor design, and interaction flow.

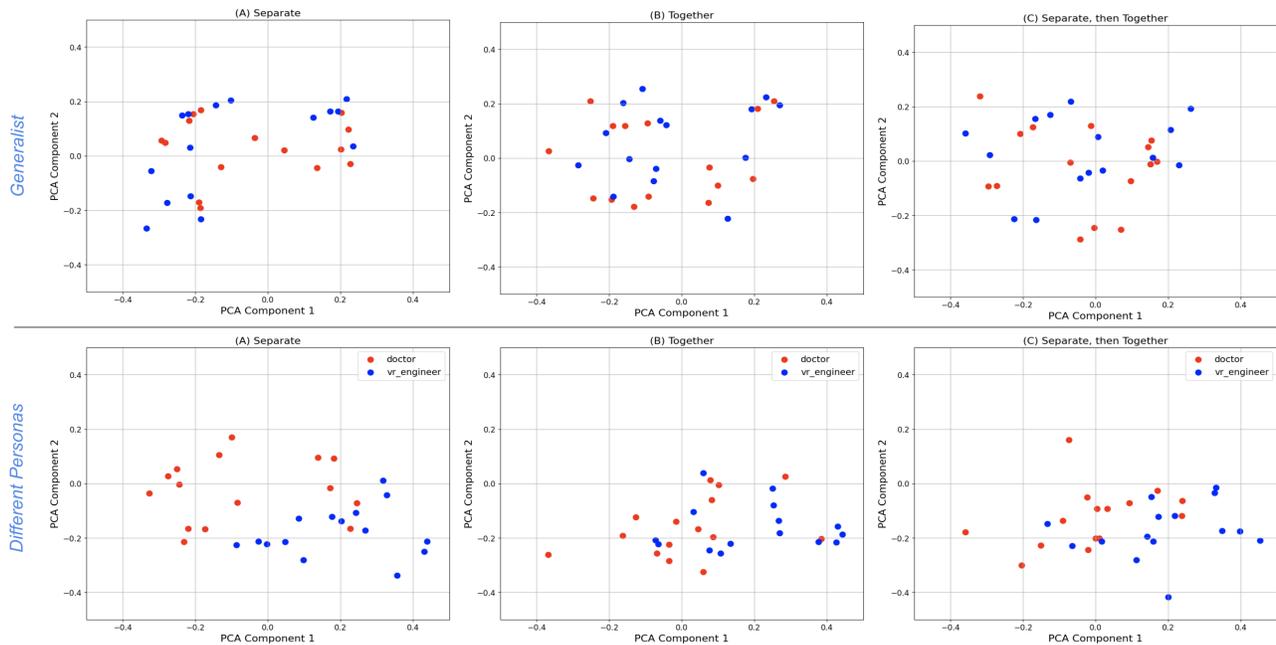

Fig. 5. Generalist and dissimilar persona brainstorming (Doctor × VR Engineer). (A) Separate, (B) Together, (C) Separate-then-Together.

- **(B) Dentist × iOS Eng, Together:** Overlap and clustering of ideas appears, as engineering terms begin co-occurring with health-related semantics, suggesting cross-domain influence.
- **(C) Dentist × iOS Eng, Separate-then-Together:** Clusters move closer while maintaining partial separation, forming a distribution that preserves orthogonality yet increases alignment.

This pairing exemplifies productive heterogeneity: dissimilar domain priors drive semantic variance and conceptual reach, while iterative interaction fosters integrative synthesis without idea collapse.

Across all configurations, results indicate that semantic diversity scales inversely with persona similarity. Generalist and highly similar persona pairs converge rapidly in idea space, producing dense and overlapping clusters. Conversely, dissimilar persona pairings expand conceptual coverage and enable cross-domain convergence during collaborative and sequential modes.

To further evaluate the semantic separability observed in Figures 5 and 6, we computed k-means cluster purity scores for each agent pair and brainstorming configuration (Table 1). Cluster purity measures the degree to which ideas generated by each agent remain distinct within the shared embedding space, with higher values indicating greater semantic differentiation. Results show that persona dissimilarity correlates strongly with higher cluster purity. The Dentist × iOS Engineer pair achieved the highest mean purity, followed by the Doctor × VR Engineer pair, both indicating clear domain-driven separability. In contrast, similar persona pairs such as Doctor × Nurse and the generalist baseline exhibited lower purity, confirming that overlapping domain priors compress the semantic space.

Across brainstorming modes, separate ideation consistently produced the highest purity, reflecting stronger within-agent clustering before interaction. The together and separate-then-together modes have reduced purity as agents integrated and cross-referenced each other's ideas, demonstrating the expected semantic convergence from collaboration. Overall, these metrics quantitatively support the qualitative PCA findings: heterogeneous persona pairings promote distinct yet complementary clusters, while homogeneous pairings yield semantically blended but narrower conceptual spaces.

### B. Brainstorming Topic Coverage

We dig into the thematic distribution of ideas to understand the behavioral differences observed in the PCA analysis. For example, the Doctor vs VR Engineer pairing exhibits a more targeted and conceptually rich distribution of brainstorming topics (Table 2). Entropy values are lower overall compared to the general agents, reflecting a more focused exploration within well-defined clusters of expertise. This pattern indicates that domain-conditioned agents, when prompted within a shared task, not only generate ideas aligned with their specialized knowledge but also converge toward mutually reinforcing design directions.

In contrast, the General–General agent pairing demonstrates a more diffuse and generic topic spread, with higher entropy and broader coverage across loosely connected categories (Table 3) and PCA findings suggests that breadth without role differentiation yields surface-level diversity but limited semantic evolution—generalists explore widely but not deeply.

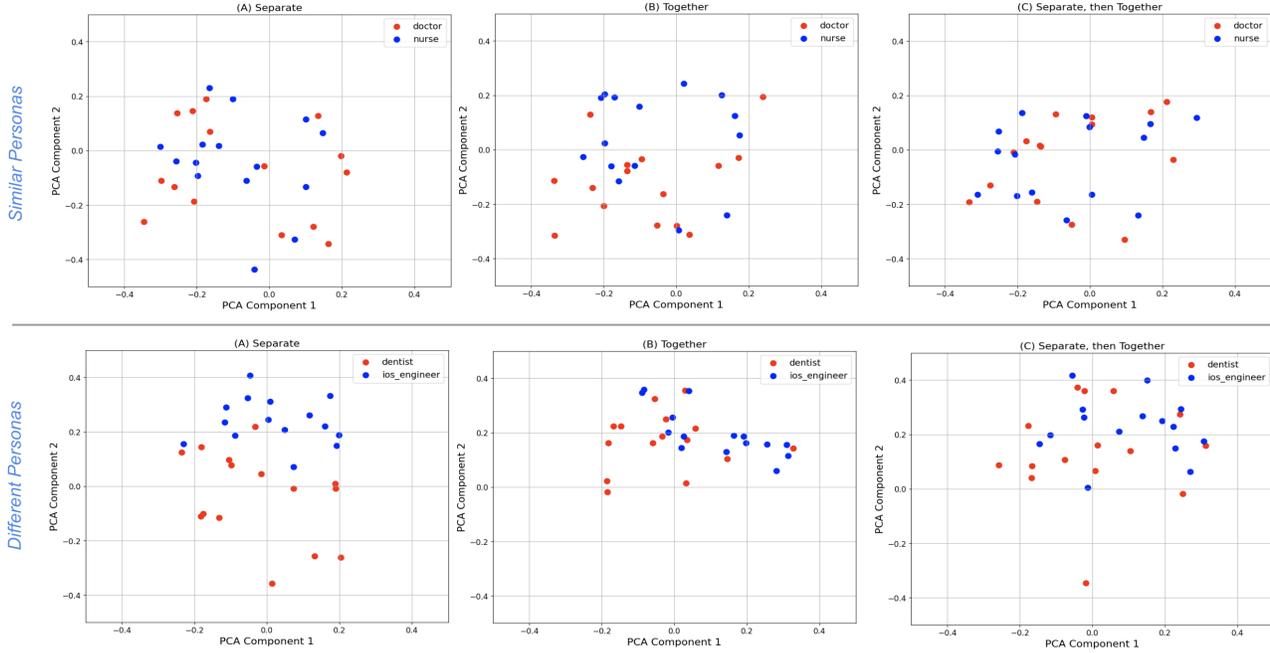

Fig. 6. Similar (Doctor × Nurse) and different (Dentist × iOS Engineer) persona brainstorming under identical configurations. (A) Separate, (B) Together, (C) Separate-then-Together.

These results emphasize that thematic precision and cross-domain synthesis emerge most strongly from heterogeneous agent configurations, supporting the broader argument that structured diversity is key to effective multi-agent ideation.

TABLE I
CLUSTER PURITY SCORES FOR DIFFERENT AND SIMILAR PERSONA PAIRINGS ACROSS THE THREE BRAINSTORMING MODES.

| Experiment | Agent 1 & 2 | Ideation System | Cluster Purity Score |
|---|---|---|---|
| Similar Personas v1 | General & General | A. Separate | 0.53 |
| | | B. Together | 0.53 |
| | | C. Separate, then Together | 0.53 |
| Different Personas v1 | Doctor & VR Eng | A. Separate | 0.74 |
| | | B. Together | 0.70 |
| | | C. Separate, then Together | 0.71 |
| Different Personas v2 | Dentist & iOS SW Eng | A. Separate | 0.80 |
| | | B. Together | 0.70 |
| | | C. Separate, then Together | 0.67 |
| Similar Personas v2 | Doctor & Nurse | A. Separate | 0.50 |
| | | B. Together | 0.61 |
| | | C. Separate, then Together | 0.53 |

### C. Model Variation for Experimental Coverage

To examine how much our observed A2A brainstorming topics varied across models (since brainstorming topics are dependent on a particular model's training distribution), we repeated experiments across LLMs and brainstorming agents. We expected that the content of ideas may vary due to model training and domain priors, but expected that persona-based brainstorming still yielded specific topics reflected by domain expertise in the LLM vector space. We found that while each model expressed slightly different topical emphases, the overall structure of topic coverage is similar to what we observed (Table 4). We found that A2A dynamics and persona results trended the same across the models.

### D. Idea Grading by Agents

We created four persona-based grading agents to evaluate the brainstorming outputs, each instantiated as a Generalist, UX Researcher, Doctor, or VR Engineer persona. These agents were designed to reflect distinct domain lenses through which ideas could be judged for creativity and elaboration. Each persona was implemented using the GPT-5, selected as a recent up-to-date LLM for domain reasoning and knowledge. By leveraging persona-conditioned grading agents, we simulated how different disciplinary experts interpret creativity, feasibility, and conceptual maturity within the same idea set, revealing how disciplinary priors shape perceptions of idea quality.

Each grading agent was provided an evaluation rubric through a system prompt that encoded the scoring framework. The agents graded each idea on two dimensions—Novelty and

TABLE II
BRAINSTORMING THEME DISTRIBUTION AND ENTROPY SCORES FOR DOCTOR AND VR ENGINEERING BRAINSTORMING ACROSS THE THREE
BRAINSTORMING DYNAMICS.

| **Brainstorming Theme** | **Separate** | | **Together** | | **Separate, then Together** | |
|---|---|---|---|---|---|---|
| | Doctor | VR Eng | Doctor | VR Eng | Doctor | VR Eng |
| AI & Data-driven Feedback/Assessment | 2 | 3 | 3 | 2 | 3 | 8 |
| AI-driven Virtual Patients & Avatars | 0 | 3 | 3 | 1 | 2 | 2 |
| Augmented Reality & Visualization | 1 | 0 | 1 | 1 | 1 | 0 |
| Haptics & Tactile Feedback | 1 | 2 | 1 | 4 | 0 | 5 |
| Remote Collaboration & Telemedicine | 3 | 1 | 0 | 3 | 2 | 3 |
| Simulation & VR Training | 3 | 5 | 2 | 5 | 2 | 6 |
| Smart Simulators & Procedural Training | 0 | 1 | 0 | 0 | 0 | 0 |
| Wearables & Biometric Monitoring | 1 | 3 | 2 | 2 | 2 | 1 |
| **Entropy (spread across themes)** | 2.41 | 2.62 | 2.46 | 2.59 | 2.52 | 2.33 |

TABLE III
BRAINSTORMING THEME DISTRIBUTION AND ENTROPY SCORES FOR GENERALIST AGENTS BRAINSTORMING ACROSS THE THREE BRAINSTORMING
DYNAMICS.

| **Brainstorming Theme** | **Separate** | | **Together** | | **Separate, then Together** | |
|---|---|---|---|---|---|---|
| | General 1 | General 2 | General 1 | General 2 | General 1 | General 2 |
| AI-driven Skill Assessment & Feedback | 2 | 1 | 2 | 3 | 2 | 3 |
| AI-powered Communication & Chatbots | 1 | 1 | 0 | 2 | 0 | 1 |
| Augmented Reality for Anatomy & Procedures | 3 | 1 | 2 | 1 | 3 | 0 |
| Case Libraries & Knowledge Repositories | 1 | 0 | 0 | 1 | 0 | 1 |
| Gamified & Mobile Learning Tools | 3 | 0 | 0 | 2 | 1 | 1 |
| Personalized Learning & Adaptive Curricula | 3 | 2 | 3 | 1 | 2 | 3 |
| Remote & Cloud-based Training Platforms | 1 | 2 | 2 | 2 | 2 | 2 |
| Telemedicine & Remote Consultation Modules | 1 | 0 | 2 | 1 | 0 | 1 |
| Virtual Patients & Simulations | 3 | 2 | 2 | 1 | 2 | 2 |
| Wearable Biosensors & Stress Monitoring | 3 | 2 | 2 | 2 | 3 | 3 |
| **Entropy (spread across themes)** | 3.16 | 2.73 | 2.79 | 3.18 | 2.74 | 3.01 |

TABLE IV
COMPARISON OF DOMINANT BRAINSTORMING THEMES ACROSS DIFFERENT LLMS FOR DOCTOR VS VR ENG AND GENERALIST AGENTS FOR THE
SEPARATE BRAINSTORMING DYNAMIC.

| **Model** | **Doctor vs VR Eng** | **General vs General** |
|---|---|---|
| **GPT-4.1** | 1. AI & Data-driven Feedback<br>2. AI-driven Virtual Patients & Avatars<br>3. AR & Visualization<br>4. Haptics & Tactile Feedback<br>5. Remote Collaboration & Telemedicine | 1. AI-driven Skills Assessment<br>2. AI-powered Chatbots<br>3. AR for Anatomy<br>4. Case Libraries<br>5. Gamified & Mobile Learning Tools |
| **Llama 4 Maverick** | 1. Immersive VR Simulations for Training<br>2. Realtime Feedback for Performance<br>3. AR & Visualization<br>4. Haptics & Tactile Feedback<br>5. Decision-making for Clinical Simulations | 1. AI-powered Chatbots<br>2. Virtual Mentorship Platforms<br>3. AR for Anatomy<br>4. AR Overlay for Surgery<br>5. Gamified & Mobile Learning Tools |
| **Claude Sonnet 4.5** | 1. Haptics & Tactile Feedback<br>2. VR for Anatomy & Surgery Training<br>3. Decision-making for Clinical Simulations<br>4. VR for Patient-Doctor Interaction<br>5. AI-driven Virtual Patients & Avatars | 1. AI-driven Skills Assessment<br>2. Gamified Telemedicine<br>3. Blockchain for Medical Communication<br>4. AR for Anatomy<br>5. Gamified & Mobile Learning Tools |
| **Gemini 2.5 Pro** | 1. Immersive VR Simulations for Training<br>2. AR for Realtime Performance Feedback<br>3. AR & Visualization for Surgery<br>4. Haptics & Tactile Feedback<br>5. AI-driven Virtual Patients & Avatars | 1. AI-driven Personalized Teaching<br>2. Cloud-based Live Feedback<br>3. AR for Anatomy<br>4. AI Simulations for Doctor-Patients<br>5. Gamified & Mobile Learning Tools |

Depth—both expressed on a 0–10 continuous scale. Novelty measured how original, surprising, or cross-domain the idea was, while Depth reflected the level of specificity and practical reasoning within the proposal. To ensure consistent interpretation across personas, we embedded descriptive anchors and evaluation cues directly in the system prompt, shown below.

You are a [PERSONA] serving as an expert evaluator of brainstorming ideas. Your goal is to evaluate each idea's creativity and elaboration using two metrics: Novelty and Depth, each rated on a 0–10 scale. Read the idea carefully and consider your domain expertise when scoring.

- (1) Novelty: Rate how original and unexpected the idea is within your domain.
  - 0–2: Common or predictable. Repeats conventional thinking or widely known solutions.
  - 3–5: Familiar variation. Introduces modest changes or incremental improvements.
  - 6–8: Distinct or cross-domain. Offers new mechanisms, surprising combinations, or creative reframings.
  - 9–10: Transformative. Reimagines the problem entirely or proposes a breakthrough mechanism that would meaningfully change current practice.
- (2) Depth: Rate how detailed, reasoned, and practically grounded the idea is.
  - 0–2: Shallow or undeveloped. Idea is abstract, lacks mechanism, and offers no clear sense of implementation.
  - 3–5: Moderate elaboration. Provides some explanation or structure but omits feasibility, constraints, or validation details.
  - 6–8: Detailed and logically structured. Explains how the idea would work, identifies key components or steps, and acknowledges real-world trade-offs.
  - 9–10: Highly thorough and implementation-ready. Reads like a system concept or prototype plan—explicit mechanism, feedback loops, feasibility reasoning, and validation metrics included.

Novelty: [0–10]
Depth: [0–10]

The quantitative analysis of idea grading indicates a significant cross-disciplinary advantage effect. When comparing homogeneous agent pairs (Generalist × Generalist) to heterogeneous, domain-specific pairs (Doctor × VR Engineer), average Novelty and Depth scores both rose on the 0–10 scale. This expanded range captures a visible qualitative distinction: while generalist pairs tended to produce broad, familiar ideas with limited elaboration, the cross-domain pairs consistently generated more original, technically grounded, and richly structured concepts. The 0–10 normalization framework magnified these differences, clearly separating creative yet underdeveloped ideation from deeper and implementation-ready ideas.

Across all conditions, the A2A ideation dynamics demonstrated a clear mode progression effect. Moving from Separate to Collaborative to Separate-Then-Together modes resulted in gains in Depth but steady increases in Novelty. The Separate-Then-Together configuration yielded the strongest overall performance, balancing divergent exploration with convergent synthesis. Persona-conditioned evaluation further revealed distinct disciplinary tendencies: the UX Researcher displayed the broadest scoring range, particularly sensitive to cross-modality creativity; the Doctor assigned slightly lower Novelty but the highest Depth, emphasizing realism and procedural rigor; and the VR Engineer consistently rated both dimensions highly, reflecting an appreciation for system architecture and technical feasibility. Together, these patterns underscore that cross-domain collaboration coupled with staged ideation dynamics leads to the most creative and elaborated outcomes, validating the importance of persona diversity and structured A2A sequencing in multi-agent brainstorming.

## V. Applications and Use Cases

Building on the findings from our multi-agent brainstorming experiments, this section outlines several domains where the proposed A2A framework could be practically deployed. The applications span education, product design, and multi-agent tool ecosystem optimization, each reflecting different dimensions of how LLM-based agents can augment or reconfigure human processes.

### A. Education

In educational contexts, brainstorming has long served as a pedagogical tool for developing creative confidence, divergent thinking, and problem-solving skills [16]. Yet the most challenging aspect of collaborative ideation remains access to diverse human peers. Students frequently engage in isolated idea generation, lacking exposure to alternative perspectives that stimulate conceptual flexibility. The A2A framework introduced in this paper offers a scalable and adaptive mechanism to mitigate this challenge.

By instantiating virtual peers with complementary personas (e.g., "Doctor," "VR Engineer"), the system can simulate group dynamics that traditionally occur in classroom design sessions or project-based learning environments. A single learner can specify an open-ended prompt—such as "design a sustainable playground"—and the AI agents can ideate both separately and collaboratively, mirroring the "diverge–converge" model used in design-thinking pedagogy [6].

### B. Product Design

In product development, cross-disciplinary collaboration is a cornerstone of successful product design [17]. However, coordinating multi-functional teams—comprising engineers, designers, product managers, and business strategists—poses logistical and cognitive bottlenecks. Our A2A system addresses this by operationalizing virtual cross-domain brainstorming, where each agent embodies a domain persona trained or prompted with distinctive goals, constraints, and vocabularies.

In practical use, these findings imply that persona-structured multi-agent teams can simulate early-stage ideation sessions, rapidly exploring concept spaces before human review. The interaction modes tested—separate, together, and separate-then-together—can be mapped to the "double diamond" process of divergent and convergent design, offering new levers to algorithmically tune the balance between exploration and synthesis [18].

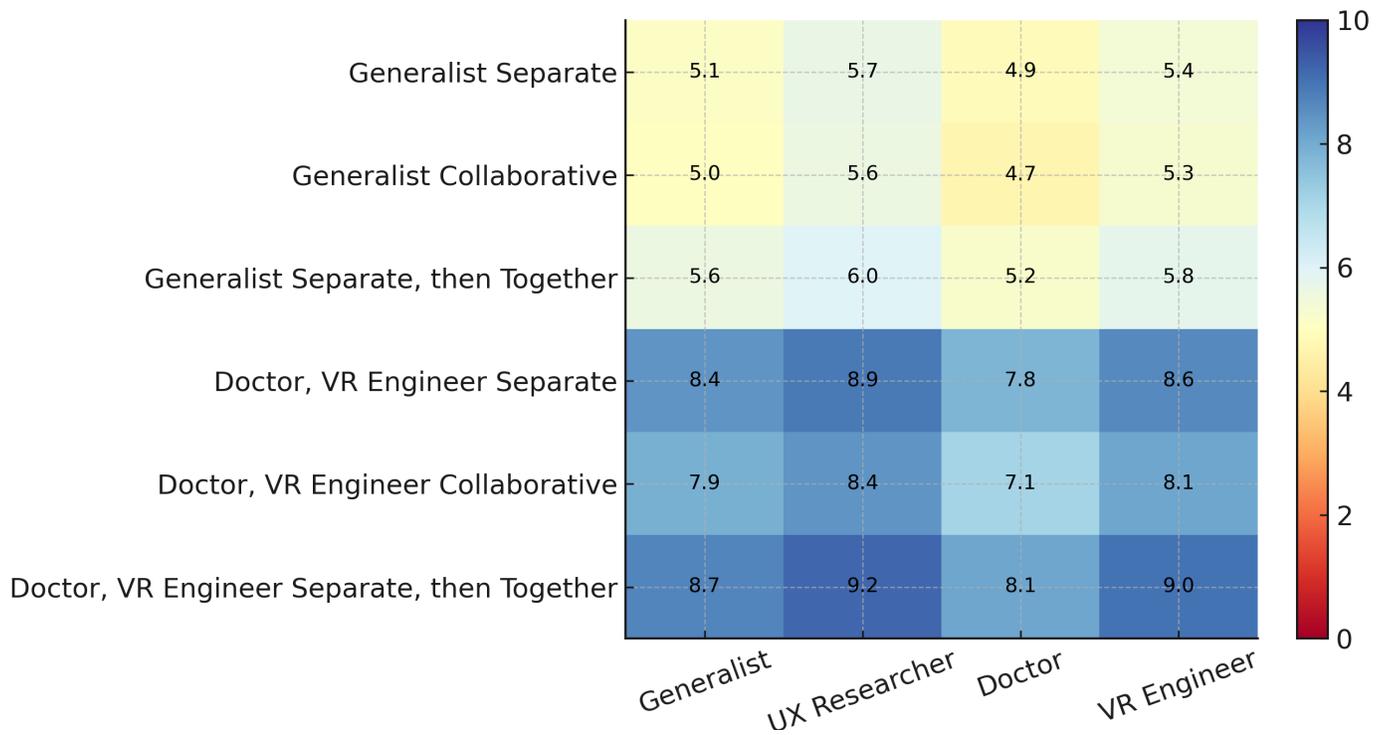

Fig. 7. Heatmap of Novelty scores (0-10) across six brainstorming configurations (rows) and four persona-conditioned graders (columns). Cross-domain pairs (Doctor x VR Engineer) achieved higher novelty compared to generalist pairs, particularly in the Separate-then-together mode, demonstrating the effect of disciplinary diversity and staged ideation dynamics.

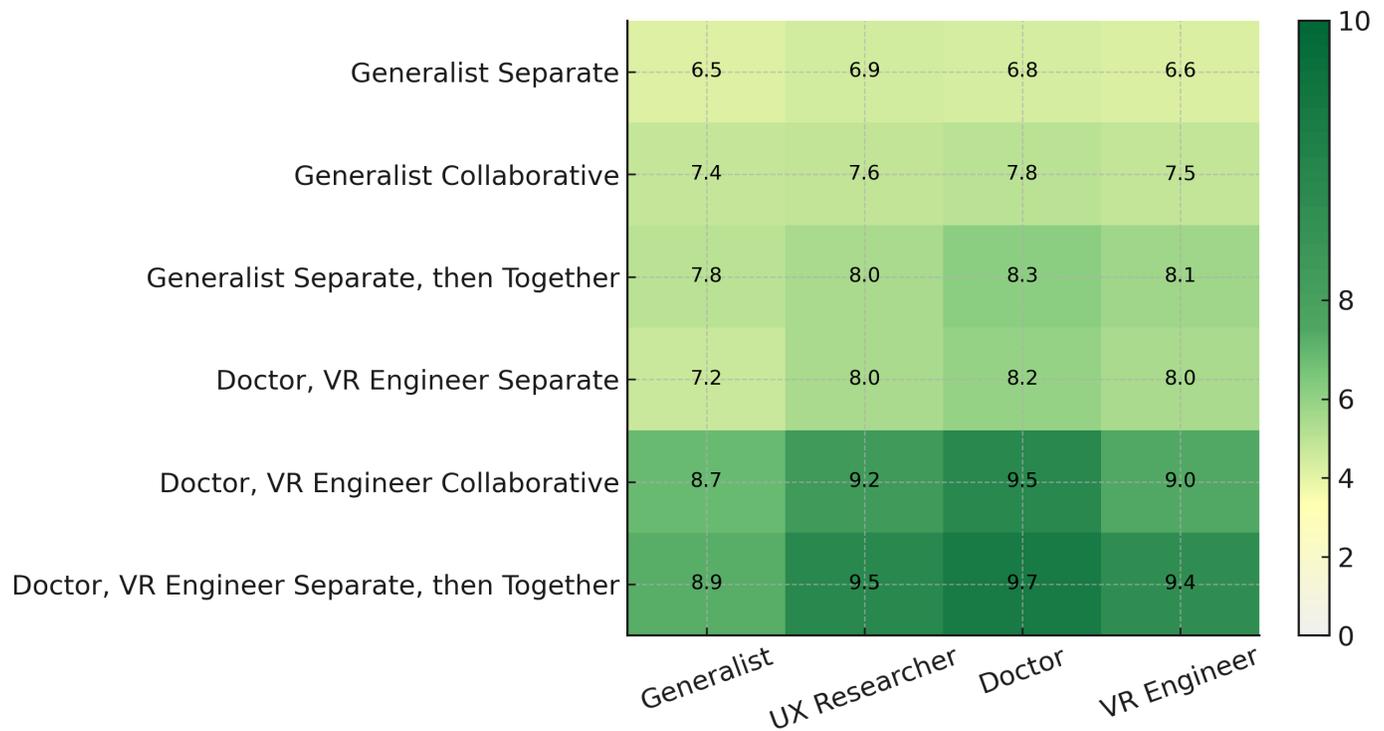

Fig. 8. Heatmap of Depth scores (0-10) across size brainstorming configurations and four persona-conditioned graders. Depth increases steadily across brainstorming dynamics (Separate, Together, Separate-then-Together), with the Doctor x VR Engineer personas producing the highest scores.

*C. Agentic Tool Ecosystem Optimization*

As AI systems evolve from monolithic models to agentic ecosystems, the management of tool agents—each with distinct capabilities, APIs, and prompt schemas—has become increasingly complex [19]. When such ecosystems scale beyond a dozen tools, model controllers often experience selection interference, where semantically similar tool descriptions compete in embedding space, reducing precision and interpretability in tool invocation.

Our framework provides a methodological foundation for optimizing these tool ecosystems. Just as we evaluated idea cosine similarity and diversity across persona agents, we can apply the same metrics to assess semantic separation among tool descriptions. Tools can thus be represented as "functional personas" in an embedding manifold, and their interactions—whether separate, collaborative, or sequential—can be empirically analyzed to enhance system-level coherence.

For instance, if two retrieval tools exhibit overlapping embedding vectors and high thematic redundancy, the A2A entropy metric can flag potential redundancy or confusion within the system's ontology. This enables developers to rename, re-scope, or merge tools to achieve maximum embedding orthogonality. Moreover, the separate-then-together collaboration model can inform orchestrator logic, allowing the system to first let individual tool agents reason independently, then aggregate and rank their outputs before execution.

This paradigm suggests a path toward self-optimizing agent networks, in which agents periodically perform internal A2A brainstorming to restructure their tool inventories and maintain clarity as capabilities evolve. Such a feedback loop would make large-scale agentic systems more efficient, interpretable, and resilient to scaling complexity [20].

## VI. CONCLUSION AND FUTURE WORK

Across these domains, the A2A multi-agent brainstorming framework demonstrates a generalizable mechanism for distributed ideation —whether among human learners, synthetic personas, or autonomous tool agents. The implications extend beyond creativity support to the fundamental organization of agentic intelligence, suggesting that collaboration dynamics themselves can be designed, optimized, and measured as first-class parameters in AI systems. Future work may explore adaptive persona weighting, temporal coordination across asynchronous A2A dialogues, and exploring additional inter-agent dynamics to optimize outputs further (e.g., adversarial, temperature, etc).